\RequirePackage{fix-cm}
\documentclass[twocolumn]{svjour3}          % twocolumn
\smartqed  % flush right qed marks, e.g. at end of proof
\usepackage{mathtools}
\usepackage{times}
\usepackage{epsfig}
\usepackage{graphicx}
\usepackage{amssymb}
\usepackage{multirow}
\usepackage{algorithm}
\usepackage[]{algpseudocode}
\usepackage{dsfont}
\usepackage{url}
\usepackage{natbib}
\usepackage{enumerate}

\usepackage{enumitem}
\usepackage{subcaption}
\usepackage{pifont}
\usepackage{blindtext}
\usepackage{morewrites}
\usepackage{color,soul}
\usepackage{algorithm}
\usepackage{algcompatible}
\usepackage{comment}
\newcommand{\xmark}{\ding{55}}%
\newcommand{\cmark}{\ding{51}}%

\makeatletter
\renewcommand{\paragraph}{%
  \@startsection{paragraph}{4}%
  {\z@}{1ex \@plus 1ex \@minus .2ex}{-1em}%
  {\normalfont\normalsize\bfseries}%
}
\makeatother

\journalname{International Journal of Computer Vision}
\begin{document}

\title{Learning from Longitudinal Face Demonstration - \\ Where Tractable Deep Modeling Meets Inverse Reinforcement Learning
}
\author{Chi Nhan Duong         \and
        Kha Gia Quach \and
        Khoa Luu \and
        T. Hoang Ngan Le \and
        Marios Savvides \and
        Tien D. Bui
}

\institute{
	Chi Nhan Duong$^1$, Kha Gia Quach$^1$, Khoa Luu$^2$, T. Hoang Ngan Le$^3$, Marios Savvides$^3$, Tien D. Bui$^1$\\
	$^1$ Department of Computer Science and Software Engineering, Concordia University, Montr\'eal, Qu\'ebec, Canada. \\
    $^2$ Computer Science and Computer Engineering Department, University of Arkansas, Fayetteville, AR.\\
    $^3$CyLab Biometrics Center and the Department of Electrical and Computer Engineering, Carnegie Mellon University, Pittsburgh, PA.\\
	\email{\tt\small \{dcnhan, kquach\}@ieee.org, khoaluu@uark.edu, thihoanl@andrew.cmu.edu, msavvid@ri.cmu.edu, bui@encs.concordia.ca}\\
}

\date{Received: date / Accepted: date}

\maketitle

\begin{abstract}
This paper presents a novel Subject-dependent Deep Aging Path (SDAP), which inherits the merits of both Generative Probabilistic Modeling and Inverse Reinforcement Learning to model the facial structures and the longitudinal face aging process of a given subject. The proposed SDAP is optimized using tractable log-likelihood objective functions with Convolutional Neural Networks (CNNs) based deep feature extraction.  Instead of applying a fixed aging development path for all input faces and subjects, SDAP is able to provide the most appropriate aging development path for individual subject that optimizes the reward aging formulation. Unlike previous methods that can take only one image as the input, SDAP further allows multiple images as inputs, i.e. all information of a subject at either the same or different ages, to produce the optimal aging path for the given subject. Finally, SDAP allows efficiently synthesizing in-the-wild aging faces. The proposed model is experimented in both tasks of face aging synthesis and cross-age face verification. The experimental results consistently show SDAP achieves the state-of-the-art performance on numerous face aging databases, i.e. FG-NET, MORPH, AginG Faces in the Wild (AGFW), and Cross-Age Celebrity Dataset (CACD). Furthermore, we also evaluate the performance of SDAP on large-scale Megaface challenge to demonstrate the advantages of the proposed solution.

\end{abstract}

\begin{figure}[t]
	\begin{center}
		\includegraphics[width=1.0\linewidth]{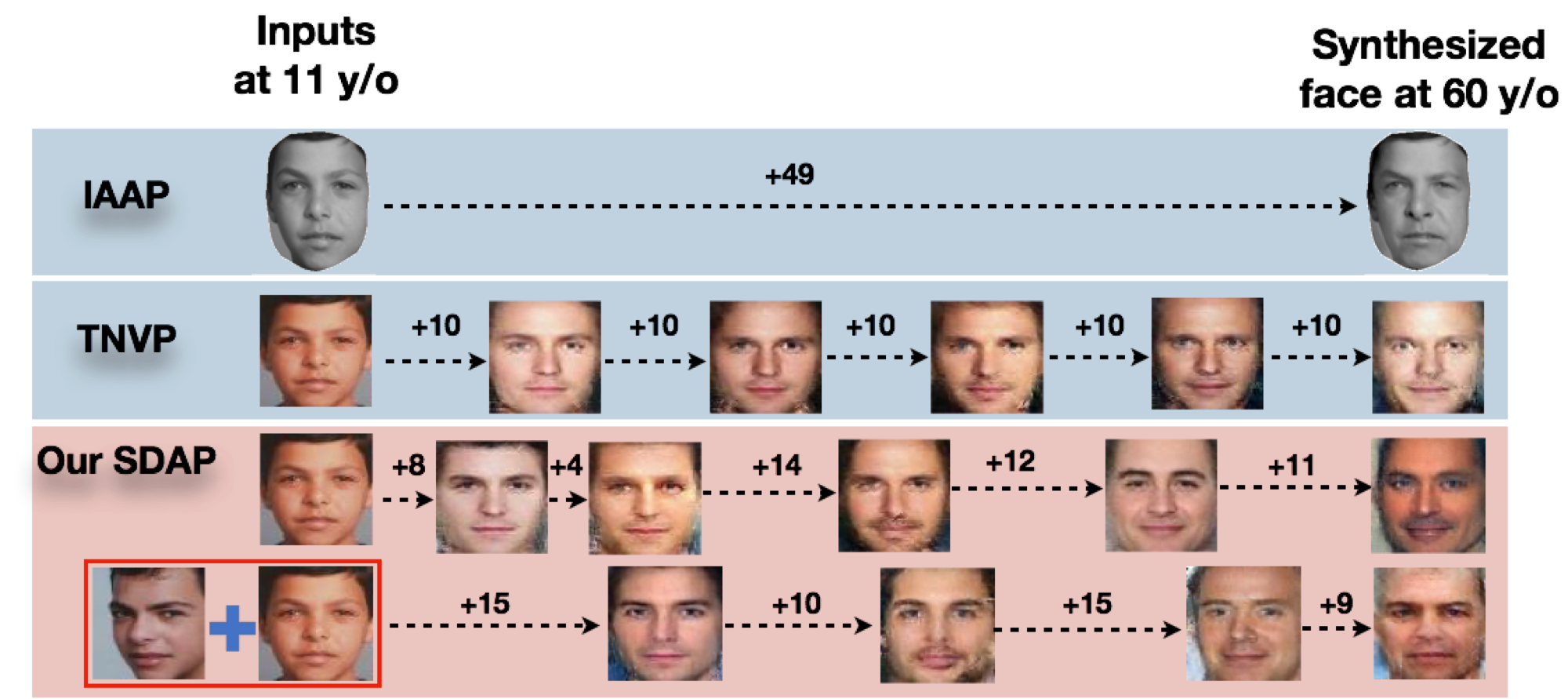}
	\end{center}
	\caption{An illustration of age progression using direct (IAAP), step-by-step (TNVP) and our SDAP approaches. Instead of using a fixed aging development path for all inputs as in IAAP (the $1^{st}$ row) and TNVP (the $2^{nd}$ row), given a face and the target age, SDAP can find the optimal aging development path (the $3^{rd}$ row) for the subject. SDAP can also collect aging cues from multiple inputs to produce more aesthetical synthesis results (the $4^{th}$ row).}
\end{figure}

\begin{table*} [!t]
	\small
	\centering
	\caption{The comparison of the properties between our SDAP approach and other age progression methods. 
		Deep learning (DL), Inverse Reinforcement Learning (IRL), Dictionary (DICT), Prototype (PROTO), Probabilistic Graphical Models (PGM), Log-likelihood (LL), Adversarial (ADV). Note that \xmark $ \: $ indicates \textit{unknown} or \textit{not directly applicable} properties.}
	\begin{tabular}{| p{4.2cm}| p{1.5cm}| p {1.5cm} | p {1.35cm}| p {1.1cm}| p {1.3cm}| p {1.1cm}| p {1.7cm}|}
		\hline
		& \textbf{Our SDAP} & TNVP \citep{Duong_2017_ICCV} & Pyramid GANs \citep{Yang_2018_CVPR} & CAAE \citep{Zhang_2017_CVPR}  & RNN \citep{wang2016recurrent} & TRBM \citep{Duong_2016_CVPR}& HFA \citep{yang2016face}  \\ \hline \hline
		\textbf{Subject-dependent Aging Path} & \cmark & \xmark & \xmark & \xmark & \xmark & \xmark & \xmark \\ \hline 
		\textbf{Multiple Input Optimization} & \cmark & \xmark & \xmark& \xmark & \xmark & \xmark & \xmark \\ \hline 
		\hline
		\textbf{Tractable} &\cmark &\cmark & \cmark & \cmark & \cmark & \xmark & \cmark  \\ \hline
		\textbf{Non-Linearity} &\cmark &\cmark & \cmark &\cmark
		& \cmark  & \cmark & \xmark  \\ \hline
		\hline
		\textbf{Model Type} & \textbf{DL + IRL} & DL & DL &DL & DL & DL & DICT \\ \hline
		\textbf{Architecture} & PGM+CNN & PGM+CNN & CNN & CNN & CNN & PGM & Bases  \\ \hline  
		\textbf{Loss Function} & LL & LL & ADV+$\ell_2$ & ADV+$\ell_2$ & $\ell_2$ & LL  & LL+$\ell_0$ \\ \hline
	\end{tabular}\label{tab:TenMethodSumm}
\end{table*}

\section{Introduction}

The problem of face aging targets on the capabilities to aesthetically \textit{synthesize} the faces of a subject at older ages, i.e. \textit{age progression}, or younger ages, i.e. \textit{age regression} or \textit{deaging}. This problem is applicable in various real-world applications from age invariant face verification, finding missing children to cosmetic studies. Indeed, face aging has raised considerable attentions in computer vision and machine learning communities recently. Several breakthroughs with numerous face aging approaches, varying from anthropology theories to deep learning structures have been presented in literature.
However, the synthesized results in these previous approaches are still far from perfect due to various challenging factors, such as heredity, living styles, etc. In addition, face aging databases used in most methods to learn the aging processes are usually limited in both number of images per subject and the covered age ranges of each subject.

Both conventional and deep learning methods usually include two directions, i.e. \textit{direct} and \textit{step-by-step aging synthesis}, in exploring the temporal face aging features from training databases. 
In the former direction, these methods \textit{directly} synthesize a face to the target age using the relationships between training images and their corresponding age labels. For example, the prototyping approaches \citep{burt1995perception, kemelmacher2014illumination,rowland1995manipulating} use  age labels to organize images into age groups and compute average faces for their prototypes. Then, the difference between source-age and target-age prototypes is applied directly to the input image to obtain the  age progressed  face at the target age.
Similarly, the Generative Adversarial Networks (GAN) approach \citep{Zhang_2017_CVPR} models the relationship between high-level representation of input faces and age labels by constructing a deep neural network generator. This generator is then incorporated with the target age labels to synthesize the outputs.
Although these kinds of models are easy to train, they are \textit{limited} in capabilities to synthesize faces \textit{much older} than the input faces of the same subject, e.g. directly from ten to 60 years old.
Indeed, the progression of a face at ten years old to the one at 60 years old in these methods usually ends up with a synthesized face using 10-year-old features plus wrinkles. 

Meanwhile, the latter approaches \citep{Duong_2017_ICCV,Duong_2016_CVPR, Shu_2015_ICCV, wang2016recurrent,yang2016face} \textit{decompose} the long-term aging process into \textit{short-term} developments and focus on the aging transform embedding between faces of two \textit{consecutive} development stages. Using learned transformation, these methods step-by-step generate progressed faces from one age group to the next until reaching the target. These modeling structures can efficiently learn the temporal information and provide more age variation even when a target age is very far from the input age of a subject. However, the main \textit{limitation} of these methods is the \textit{lack} of longitudinal face aging databases. The longest training sequence usually contains only three or four images per subject. 

\paragraph{\textbf{Limitations of previous approaches.}}
In either directions, i.e. direct or step-by-step aging synthesis, the aging approach falls in, 
these previous approaches still suffer from many challenging factors and remain with lots of limitations. Table \ref{tab:TenMethodSumm} compares the properties of different aging approaches.
\begin{itemize}
	\item \textbf{Non-linearity.} Since human aging is a complicated and highly nonlinear process, the linear models mostly used in conventional methods, i.e. prototype, AAMs-based and 3DMM-based approaches, are unable to efficiently interpret the aging variations and the quality of their synthesized results is very limited. 
	\item \textbf{Loss function of deep structure.} The use of a fixed reconstruction loss function, i.e. $\ell_2$-norm, in the proposed deep structures \citep{wang2016recurrent,yang2016face} usually produces blurry synthesis results. 
	\item \textbf{Tractability.} Exploiting the advantages of probabilistic graphical models has introduced a potential direction for deep model design and produced prominent synthesized results for the age progression task \citep{Duong_2016_CVPR}.
	\item \textbf{Data usability.} Even though a subject in training/testing set has multiple images at the same age, there is \textit{only one} image used to learn/synthesize in these methods. The other images are usually wastefully \textit{ignored}. In addition, the aging transformation embedding in these approaches is only able to proceed on images from \textit{two} age groups.
	\item \textbf{Fixed aging development path.} The learned aging development path is identically applied for all subjects which is not true in reality. Instead, each subject should have his/her own aging development.
\end{itemize}

\paragraph{\textbf{Contributions of this work.}} 

The paper presents a novel \textit{Subject-dependent Deep Aging Path (SDAP)} model to face age progression, which is an extension of our previous work \citep{Duong_2017_ICCV}.
In that work,  TNVP structure is proposed to embed the pairwise transformations between two consecutive age groups. In this work, the SDAP structure is introduced to further enhance the capability to discover \textit{the optimal} aging  development path for \textit{each} individual. This goal can be done by embedding the transformation over the whole aging sequence of a subject under an IRL framework.
Our contributions can be summarized as follows.

\begin{enumerate}
\item The aging transformation embedding is designed using (1) a \textit{tractable log-likelihood} density estimation with (2) Convolution Neural Network (CNN) structures and (3) an \textit{age controller} to indicate the amount of aging changes for synthesis. Thus, the proposed SDAP is able to provide a \textit{smoother synthesis} across faces and maximize the \textit{usability} of aging data, i.e. all images of a subject in different or the same ages are utilized.
\item Unlike most previous methods, our proposed SDAP model further enhances the capability to find \textit{the optimal aging  development path} for individual. This goal can be done by embedding the transformation over the whole aging sequence of a subject under an IRL framework. 
\item Instead of using pre-defined or add-hoc aging reward and objective functions as in most previous work, our proposed approach allows the algorithm to automatically come up with the \textit{optimal objective formulation and parameters} via a data driven strategy in training.

\end{enumerate}

We believe that this is the first work that designs an IRL framework to model the longitudinal face aging.

\section{Related work}

This section reviews recent methods in facial aging including age estimation and age progression problems. 

\paragraph{\textbf{Age Progression.}}
Face age progression methods can be technically classified into four categories, i.e. modeling, reconstruction, prototyping, and deep learning-based methods.

\textit{Modeling-based aging} is one of the earliest categories presented face age progression. These methods usually model both facial shapes and textures using a set of parameters, and learn the face aging process via aging functions.
\citep{patterson2006automatic} and \citep{lanitis2002toward} employed a set of Active Appearance Models (AAMs) parameters with four aging functions to model both the general and the specific aging processes.
\cite{luu2009Automatic} combined familial facial cues to the process of face age progression.
\cite{geng2007automatic} presented  AGing pattErn Subspace (AGES) method to construct a subspace for aging patterns as a chronological sequence of face images. 
\cite{tsai2014human} then enhanced AGES using guidance faces corresponding to the subject's characteristics to produce more stable results. Texture synthesis was also combined in the later stage to produce better facial details.
\cite{suo2010compositional, suo2012concatenational} introduced the three-layer And-Or Graph (AOG) of smaller parts, i.e. eyes, nose, mouth, etc., to model a face. Then, the face aging process was learned for each part using a Markov chain.

\textit{Reconstruction-based aging methods} model aging faces by unifying the aging basis in each group. 
\cite{yang2016face} represented person-specific and age-specific factors independently using sparse representation Hidden Factor Analysis  (HFA).
\cite{Shu_2015_ICCV} presented the aging coupled dictionaries (CDL) to model personalized aging patterns by preserving personalized facial features.

\textit{Prototyping-based aging methods} employ the age prototypes to produce new face images. 
The average faces of all age groups are used as the prototypes \citep{rowland1995manipulating}. Then, input face image can be progressed to the target age by incorporating the differences between the prototypes of two age groups \citep{burt1995perception}.
\cite{kemelmacher2014illumination} presented a method to construct high quality average prototypes from a large-scale set of images. The subspace alignment and illumination normalization were also included in this system. Aging patterns across genders and ethnicity were also investigated in \citep{guo2014study}.

\textit{Deep learning-based approaches} have recently  achieved considerable results in face age progression using the power of deep learning. 
\cite{Duong_2016_CVPR} introduced 
Temporal Restricted Boltzmann Machines (TRBM) to represent the non-linear aging process with geometry constraints and spatial RBMs to model a sequence of reference faces and wrinkles of adult faces.
\cite{wang2016recurrent} approximated aging sequences using a Recurrent Neural Networks (RNN) with two-layer Gated Recurrent Unit (GRU). 
Recently, the structure of Conditional Adversarial Autoencoder (CAAE) is also applied to synthesize aged images in \citep{antipov2017face}. 
\cite{Duong_2017_ICCV} proposed a novel generative probabilistic model, named Temporal Non-Volume Preserving (TNVP) transformation, to model a long-term facial aging process as a sequence of short-term stages. 
The Conditional Generative Adversarial Networks were also incorporated with perceptual loss in \citep{wang2018face_aging}.
Similarly, \cite{Yang_2018_CVPR} integrated an age estimator with three labels, i.e. young /senior for real faces, and synthesized face, together with an $\ell_2$ loss between face descriptors of input and synthesized faces into a Generative Adversarial Network structure for face age progression. 

\paragraph{\textbf{Age Estimation.}}
Existing Age estimation approaches usually consist of two components: an age feature extraction module and an age prediction module.
More detailed surveys on age estimation can be found in \citep{fu2010age,angulu2018age}. 
One of the early works by \cite{kwon1999age} measured the facial shape changes using geometric of key features to classify faces into three age groups, i.e. baby, young adult and senior adult. 
Later, Active Appearance Models (AAM) \citep{cootes2001active} were adopted in several works to unify the facial shape and texture (i.e. wrinkles) features for age estimation \citep{lanitis2004comparing, luu2009age, choi2011age, zhang2010multi, chang2011ordinal, duong2011fine, Xu_IJCB2011, Xu_TIP2015, le2015facial, Luu_CAI2011, Chen_FG2011}.

Meanwhile, \cite{geng2007automatic,geng2008facial}, proposed to learn the aging pattern subspace (AGES) from datasets to model aging process over the years. Similarly, \cite{guo2008image} introduced manifold learning based approach together with locally adjusted regressor for age prediction. In addition, \cite{guo2012study} studied the relationship between facial age and facial expression for more robust estimator. Instead of directly predicting the age from facial images having some expressions, they learn the correlation between two expressions (e.g. happy vs. neutral) to map the face from one expression to another and then predict the age on the “transformed” face.
Some other approaches focused on feature extraction process with hand-crafted features such as biologically inspired features (BIF) \citep{guo2009human}, spatially flexible patch (SFP) \citep{yan2008extracting}, synchronized submanifold embedding (SME) \citep{yan2009synchronized}, Gabor filters and local binary pattern (LBP) features \citep{choi2011age}. Then linear classifiers/regressors such as Support Vector Machine (SVM), Support Vector Regression (SVR) or locally adjusted robust regression (LARR) were adopted for age estimation.

Recently, with the success of deep convolution neural networks (CNNs) in many face-related tasks including face detection, face alignment and face verification, a wide range of end-to-end CNN-based age estimation methods have been proposed to address this challenging problem. Different deep network structures have been exploited in these works such as \textit{Multi-scale CNN} \citep{yi2014age}, \textit{Deep Appearance Models} (DAMs) \citep{nhan2015beyond}, \textit{VGGFace} \citep{yang2015deep, Rothe-IJCV-2016}, \textit{Ordinal-CNN} \citep{niu2016ordinal}, \textit{Deep Regression Forests} \citep{shen2018deep}.

In particular, \cite{yi2014age} proposed to use CNNs with multi-scale analysis, local aligned face patch, and facial symmetry strategies from traditional approaches.
\cite{nhan2015beyond} developed a non-linear model combining shape and texture information called Deep Appearance Models which was inspired from the traditional AAMs for age estimation and face modeling.
\cite{niu2016ordinal} considered age estimation as an ordinal regression problem which is converted into multiple binary classifications.
Similarly, \cite{chen2017using} also proposed a set of CNNs with binary output and then the final prediction is obtained via aggregation.
\cite{agustsson2017anchored} proposed an anchored regression network consisting of multiple linear regressors assigned to an anchor point to linearize the regression problem.
\cite{shen2018deep} introduced Deep Regression Forests (DRFs) for age estimation. In DRFs, a CNN with FC layer is connected to inner nodes in DRFs, and then leaf node distribution output in each tree is combined to provide the final predictions.
\cite{pan2018mean} proposed to learn CNNs by jointly optimizing mean-variance loss and softmax loss.
\cite{li2018deep} presented a Deep Cross-Population (DCP) model to transfer the model learned from one set with large number of labeled data to one population having small number of labeled data.
In addition, Deep Age Distribution Learning (DADL) via CNNs approaches were proposed in \citep{yang2015deep} and \cite{huo2016deep} to handle apparent age estimation problem where it predicted age distribution instead of the exact age. \cite{Rothe-IJCV-2016} proposed an ensemble of 20 VGG-liked networks called Deep EXpectation (DEX) of apparent age with softmax outputs of 101 age classes. 

Other approaches target on dealing with other facial attributes, e.g. expression, gender, race, etc.  More recently, \cite{lou2018expression} proposed to use a graphical model to jointly learn the relationship between the age and expression causing changes in facial appearance. Their method makes age estimation problem invariant to expression changes. 
\cite{wang2017deep} and \cite{han2018heterogeneous} proposed multi-task learning approaches for jointly and effectively estimating multiple facial attributes including age, gender, race and others. 

\begin{figure*}[t]
	\begin{center}
		\includegraphics[width=17.5cm]{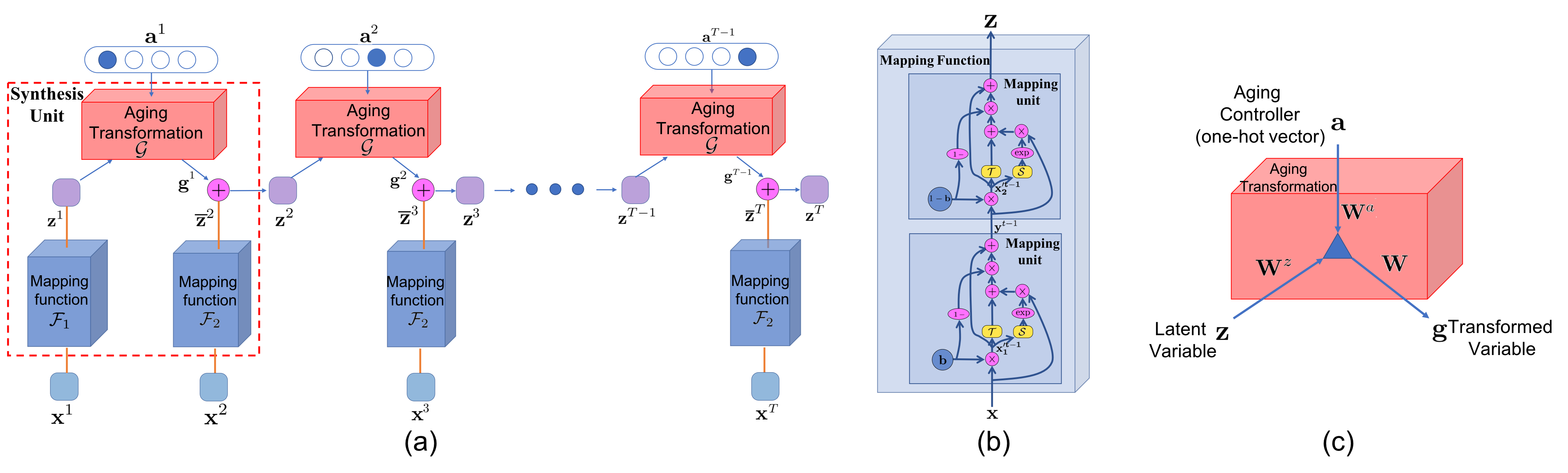}
	\end{center}
	\caption{The structures of (a) a Synthesis Component as a composition of Synthesis Units; (b) a mapping function with two mapping units; and (c) the aging transformation component. During synthesizing process, the value for aging controller $\mathbf{a}$ at each step is predicted by a Policy learned through an Inverse Reinforcement Learning framework.}
	\label{fig:AgeProgressionFramework_A}
\end{figure*}

\section{Our Proposed SDAP}
TNVP structure has provided an efficient model to capture the pairwise transformation between faces of consecutive age groups \citep{Duong_2017_ICCV}. However, it still has some limitations. Firstly, the TNVP  mainly focuses on the pairwise relationship rather than \textit{the long-term relationship presented in an aging sequence}. Secondly,  \textit{capability of applying different development paths for different subjects} is still absent. In reality, each subject should have his/her own aging development progress because each person ages differently. In this section, we introduce a more flexible structure, named Subject-dependent Deep Aging Path (SDAP), with an additional component, i.e. an age controller. This age controller provides the capability of defining how much age variation should be added during synthesis. This architecture, therefore, benefits both training stages, i.e. by maximizing the usability of training aging data, and testing stage, i.e. becoming more flexible to adopt different aging path to different subjects according their features. Moreover, instead of only learning from image pairs of a subject in two consecutive age groups, SDAP has the capability of embedding the aging transformation from longer aging sequences of that subject which efficiently reflects the long-term aging development of the subject. 
We also show that goal can be achieved under an Inverse Reinforcement Learning (IRL) framework. The structure of this section is as follows:
We first present our novel approach to model the facial structures in Subsection \ref{subsec:aging_model}. Then, our IRL learning approach to the longitudinal face aging modeling is detailed in Subsection \ref{subsec:IRL_model}.

\subsection{Aging Embedding with Age Controller}
\label{subsec:aging_model}
The proposed architecture consists of three main components, i.e. (1) latent space mapping, (2) aging transformation, and (3) age controller. Our age controller provides the capability of defining how much age variation should be added during synthesis. Using this structure, our model is flexible to aging in  different ways corresponding to the input faces. Moreover, it also helps to maximize the usability of training aging data.

\paragraph{Structures and Variable Relationship Modeling:}
Our graphical model (Fig. \ref{fig:AgeProgressionFramework_A}) consists of three sets of variables: observed variables $\{\mathbf{x}^{t-1}, \mathbf{x}^t\} \in \mathcal{I}$ encoding the textures of face images in the image domain $\mathcal{I} \subset \mathbb{R}^D$ at two stages $t-1$ and $t$; their corresponding latent variables $\{\mathbf{z}^{t-1}, \mathbf{z}^t\} \in \mathcal{Z}$ in the latent space $\mathcal{Z}$; and an aging controller variable $\mathbf{a}^{t-1} \in \mathcal{A} \subset \mathbb{R}^{N_a}$. The aging controller $\mathbf{a}^{t-1}$ is represented as a one-hot vector indicating how many years old the progression process should perform on $\mathbf{x}^{t-1}$. 
The bijection functions $\mathcal{F}_1,\mathcal{F}_2$, mapping from the observation space to the latent space, and the aging transformation $\mathcal{G}$ are defined as in Eqn. \eqref{eqn:F1F2I}.
\small
\begin{eqnarray}
\begin{alignedat}{3}
\mathcal{F}_1,\mathcal{F}_2&: &&\mathcal{I} &&\rightarrow \mathcal{Z}\\
& &&\mathbf{x}^{t-1} &&\mapsto \mathbf{z}^{t-1} = \mathcal{F}_1 (\mathbf{x}^{t-1};\boldsymbol{\theta}_1)\\
& &&\mathbf{x}^{t} &&\mapsto \mathbf{\bar{z}}^{t} = \mathcal{F}_2 (\mathbf{x}^{t};\boldsymbol{\theta}_2)\\
\mathcal{G}&: &&\mathcal{Z}, \mathcal{A} &&\rightarrow \mathcal{Z}\\
& &&\mathbf{z}^{t-1}, \mathbf{a}^{t-1} &&\mapsto \mathbf{g}^t = \mathcal{G} (\mathbf{z}^{t-1}, \mathbf{a}^{t-1};\boldsymbol{\theta}_3)\\
\end{alignedat}
\label{eqn:F1F2I}
\end{eqnarray}
\normalsize
where $\boldsymbol{\theta} = \{\boldsymbol{\theta}_1, \boldsymbol{\theta}_2, \boldsymbol{\theta}_3\}$ denotes the set of parameters of $\mathcal{F}_1$, $\mathcal{F}_2$, $\mathcal{G}$, respectively. Notice that in SDAP, the structure of bijection functions is adopted from TNVP architecture.
Then, the relationship between latent variables $\mathbf{z}^t,\mathbf{z}^{t-1}$ and $\mathbf{a}^{t-1}$ is computed as $\mathbf{z}^t = \mathbf{g}^t + \mathbf{\bar{z}}^t$.

The interactions between latent variables $\{\mathbf{z}^{t-1}, \mathbf{z}^t\}$ and the aging controller variable $\mathbf{a}^{t-1}$ are 3-way multiplicative. They can be mathematically encoded as in Eqn. \eqref{eqn::threeways}.
\small
\begin{equation}
\label{eqn::threeways}
g_i^t = \sum_{j,k} \hat{w}_{ijk} z_j^{t-1}a^{t-1}_k + b_i
\end{equation}
\normalsize
where $\mathbf{\hat{W}} \in \mathbb{R}^{D \times D \times N_a}$ is a 3-way tensor weight matrix and $\mathbf{b}$ is the bias of these connections. 
Eqn. \eqref{eqn::threeways} enables two important properties in the architecture. 
First, since $\mathbf{a}^{t-1}$ is an one-hot vector, different controllers will enable different sets of weights to be used. Thus, it allows controlling the amount of aging information to be embedded to the aging process. 
Second, given the age controller, the model is able to use all images of a subject to enhance its performance. 

\begin{figure}[t]
	\centering \includegraphics[width=8.6cm]{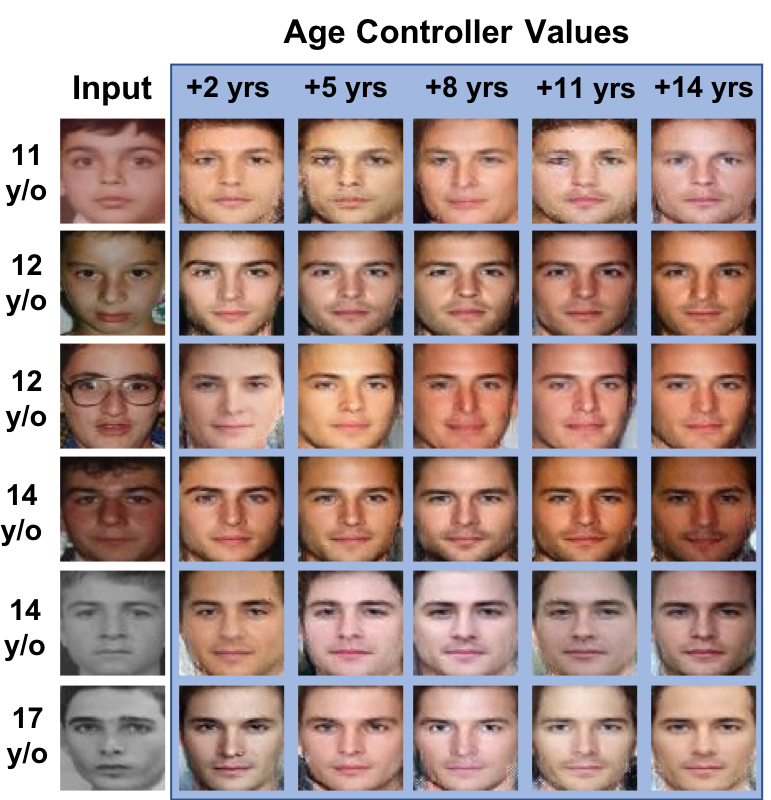}
	\caption{Synthesized results with different values of aging controller. Given an input image, by varying the values of aging controller, different age-progressed images can be obtained. Notice that the age controller can help to efficiently control the amount to aging features to be embedded while maintaining other variations between synthesized faces. \textbf{Best viewed in color.}}
\label{fig:SDAP_DifferentAction}
\end{figure} 

In practice, the large number of parameters of the 3-way tensor matrix may have negative effects to the scalability of the model. Thus, $\mathbf{\hat{W}}$ can be further factorized into three matrices $\mathbf{W}^a \in \mathbb{R}^{f \times N_a}$, $\mathbf{W}^z \in \mathbb{R}^{f \times D}$, and $\mathbf{W} \in \mathbb{R}^{D \times f}$ with $f$ factors by adopting \citep{taylor2009factored} as in Eqn. \eqref{eqn::factorize}.
\small
\begin{equation}
\mathbf{g}^t = \mathbf{W} (\mathbf{W}^z \mathbf{z}^{t-1} \odot \mathbf{W}^a\mathbf{a}^{t-1}) + \mathbf{b}
\label{eqn::factorize}
\end{equation}
\normalsize
where $\odot$ stands for the Hadamard product.

\paragraph{The Log-likelihood:}
Given a face $\mathbf{x}^{t-1}$ in the age group $t-1$, the probability density function can be formulated as,
\small
\begin{eqnarray}
\begin{split} \label{eqn:likelihood}
p_{X^t}(\mathbf{x}^t|\mathbf{x}^{t-1}, \mathbf{a}^{t-1};\boldsymbol{\theta})&=p_{X^t}(\mathbf{x}^t|\mathbf{z}^{t-1}, \mathbf{a}^{t-1};\boldsymbol{\theta})\\
&=p_{Z^t}(\mathbf{z}^t|\mathbf{z}^{t-1}, \mathbf{a}^{t-1};\boldsymbol{\theta})\left|\frac{\partial \mathbf{z}^{t}}{\partial \mathbf{x}^t}\right|\\
&=\frac{p_{Z^t,Z^{t-1}}(\mathbf{z}^t,\mathbf{z}^{t-1}|\mathbf{a}^{t-1},\boldsymbol{\theta})}{p_{Z^{t-1}}(\mathbf{z}^{t-1};\boldsymbol{\theta})}\left|\frac{\partial \mathbf{z}^{t}}{\partial \mathbf{x}^t}\right|
\end{split}
\end{eqnarray} 
\normalsize
where $p_{X^t}(\mathbf{x}^t|\mathbf{x}^{t-1},\mathbf{a}^{t-1};\theta)$ and $p_{Z^t}(\mathbf{z}^t|\mathbf{z}^{t-1},\mathbf{a}^{t-1};\theta)$ are the distribution of $\mathbf{x}^t$ conditional on $\mathbf{x}^{t-1}$ and the distribution of $\mathbf{z}^t$ conditional on $\mathbf{z}^{t-1}$, respectively.
Then, the log-likelihood can be computed as follows:
\small
\begin{equation}
\begin{split} \nonumber
\log p_{X^t}(\mathbf{x}^t|\mathbf{x}^{t-1}, \mathbf{a}^{t-1};\boldsymbol{\theta}) = &\log p_{Z^t,Z^{t-1}}(\mathbf{z}^t,\mathbf{z}^{t-1}|\mathbf{a}^{t-1},\boldsymbol{\theta}) \\
&- \log p_{Z^{t-1}}(\mathbf{z}^{t-1};\boldsymbol{\theta}) + \log \left|\frac{\partial \mathbf{z}^{t}}{\partial \mathbf{x}^t}\right| 
\end{split}
\end{equation}
\normalsize
\paragraph{The Joint Distributions:}
In order to model the aging transformation flow, the Gaussian distribution is presented as the prior distribution $p_{Z}$ for the latent space.
After mapping to the latent space, the age controller variables are also constrained as a Gaussian distribution. In particular, let $\mathbf{z}_a^{t-1} = \mathbf{W}^a \mathbf{a}^{t-1}$ represent the latent variables of $\mathbf{a}^{t-1}$. The latent variables $\{\mathbf{z}^{t-1}, \mathbf{\bar{z}}^{t}, \mathbf{\mathbf{z}}^{t-1}_a\}$ distribute as Gaussians with means $\{\boldsymbol{\mu}^{t-1}, \boldsymbol{\bar{\mu}}^{t},\boldsymbol{\mu}_a^{t-1}\}$ and covariances $\{\Sigma^{t-1},\bar{\Sigma}^{t}, \Sigma^{t-1}_a\}$ respectively. Then, the latent 
$\mathbf{z}^t$ is as,
\small
\begin{eqnarray}
\begin{alignedat}{3}
\mathbf{\mathbf{z}}^{t} &\sim \mathcal{N} &&\left(  \boldsymbol{\mu}^t,\Sigma^{t}\right)\\
\boldsymbol{\mu}^t &= \mathbf{W} &&\left(  \mathbf{W}^z\boldsymbol{\mu}^{t-1} \odot \boldsymbol{\mu}^{t-1}_a\right) + \mathbf{b} + \boldsymbol{\bar{\mu}}^t\\
\Sigma^{t} &= \mathbf{W} &&[\mathbf{W}^z \Sigma^{t-1} {\mathbf{W}^z}^{\intercal} \odot \Sigma^{t-1}_a \\
& &&- (\mathbf{W}^z \boldsymbol{\mu}^{t-1})(\mathbf{W}^z \boldsymbol{\mu}^{t-1})^\intercal \odot \boldsymbol{\mu}^{t-1}_a {\boldsymbol{\mu}^{t-1}_a}^\intercal] \mathbf{W}^\intercal
\end{alignedat}
\end{eqnarray}
\normalsize
Since the connection between $\mathbf{z}^{t-1}$ and $\mathbf{z}^t$ embeds the relationship between variables of different Gaussian distributions, we further assume that their joint distribution is also a Gaussian. 
Then, the joint distribution $p_{Z^t, Z^{t-1}}(\mathbf{z}^t,\mathbf{z}^{t-1}|\mathbf{a}^{t-1}; \theta)$ can be computed as follows.
\small
\begin{eqnarray} \nonumber
\begin{split} 
\label{eqn:JointDistribution}
p_{Z^t, Z^{t-1}}(\mathbf{z}^t,\mathbf{z}^{t-1}|\mathbf{a}^{t-1}; \theta)
&\sim \mathcal{N} 
\left( 
\begin{bmatrix}
\boldsymbol{\mu}^t\\
\boldsymbol{\mu}^{t-1}
\end{bmatrix}
,
\begin{bmatrix}
\Sigma^t & \Sigma^{t,t-1} \\
\Sigma^{t-1,t} & \Sigma^{t-1}
\end{bmatrix}
\right) \\
\Sigma^{t,t-1}&=\mathbf{W}(\mathbf{W}^z\Sigma^{t-1} \odot \boldsymbol{\mu}^{t-1}_a \mathbf{1}^\intercal)\\
\Sigma^{t-1,t}&=(\mathbf{1} {\boldsymbol{\mu}^{t-1}_a}^\intercal \odot \Sigma^{t-1} \mathbf{W}^{z\intercal}) \mathbf{W}^\intercal
\end{split}
\end{eqnarray}
\normalsize
where $\mathbf{1} \in \mathbb{R}^D$ is an all-ones vector.

\paragraph{The Objective Function:} 
The parameter $\boldsymbol{\theta}$ of the model is optimized to maximize the log-likelihood as in Eqn. \eqref{eqn:Jointargmax}.
\begin{equation}
\begin{split}
\boldsymbol{\theta}^* = &\arg \max_{\boldsymbol{\theta}}  \log p_{X^t}(\mathbf{x}^t|\mathbf{x}^{t-1}, \mathbf{a}^{t-1}; \boldsymbol{\theta})\\
&\text{s.t. } \mathbf{z}^{t-1}_a=\mathbf{W}^a \mathbf{a}^{t-1} \text{distributes as a Gaussian}
\end{split}
\label{eqn:Jointargmax}
\end{equation}
This constraint is then incorporated to the objective function
\small
\begin{equation} 
\begin{split} \nonumber
\boldsymbol{\theta}^* = \arg \max_{\boldsymbol{\theta}}  \log p_{X^t}(\mathbf{x}^t|\mathbf{x}^{t-1}, \mathbf{a}^{t-1}; \boldsymbol{\theta}) + l(\boldsymbol{\mu}^{t-1}_a, \Sigma^{t-1}_a; \mathbf{a}^{t-1})
\end{split}
\end{equation}
\normalsize
where $l(\boldsymbol{\mu}^{t-1}_a, \Sigma^{t-1}_a; \mathbf{a}^{t-1})$ is the log-likelihood function of $\mathbf{a}^{t-1}$ given mean $\boldsymbol{\mu}^{t-1}_a$ and covariance $\Sigma^{t-1}_a$.
\subsection{IRL Learning from Aging Sequence}
\label{subsec:IRL_model}

\begin{figure*}[t]
	\begin{center}
		\includegraphics[width=16cm]{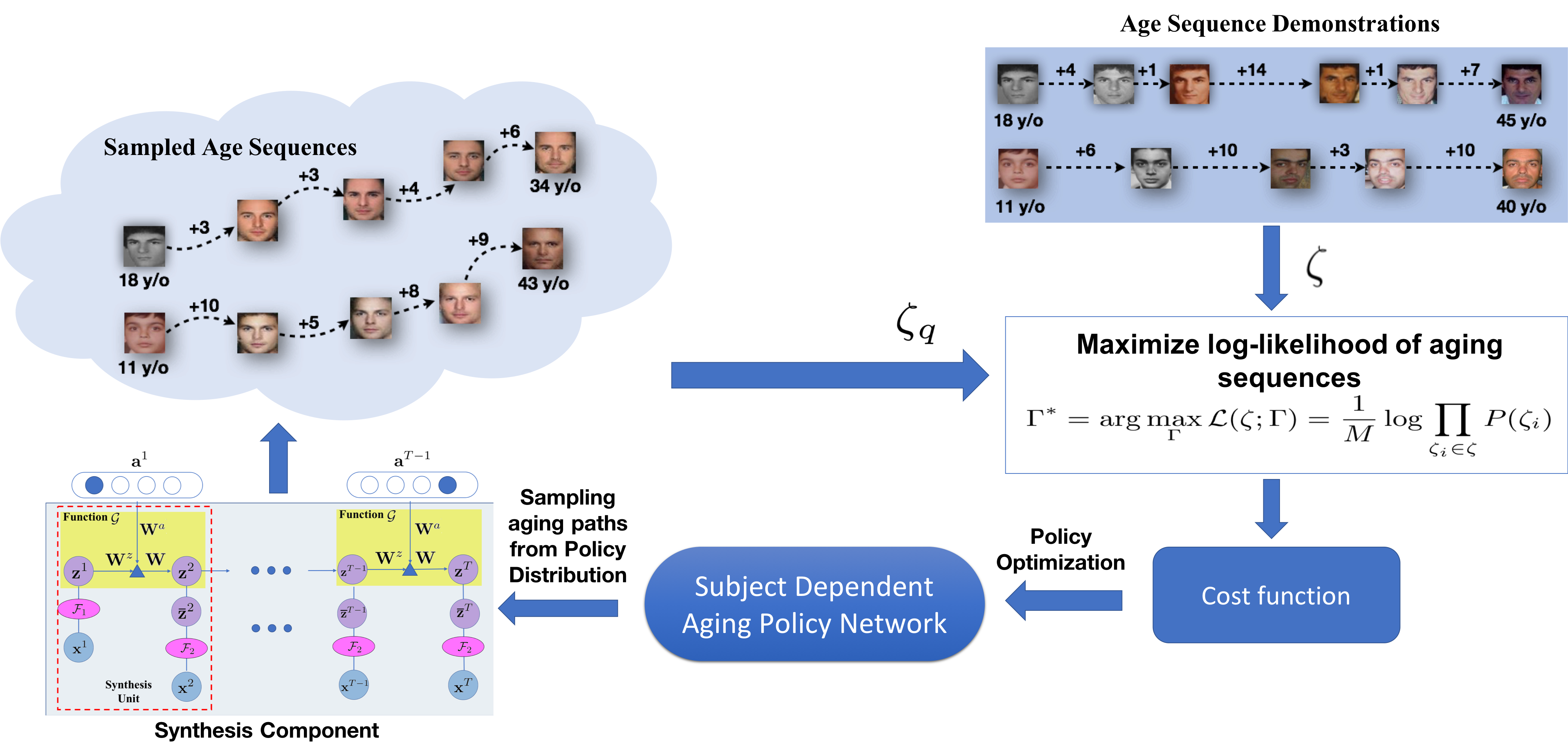}
	\end{center}
	\caption{The Subject-Dependent Aging Policy Learning Framework. Given the Age Sequence Demonstrations, the cost function is learned by maximizing the log-likelihood of these sequences. Its output is then can be used to optimize the Subject Dependent Aging Policy Network which later is able to predict the most appropriate aging path for each subject.}
	\label{fig:AgeProgressionFramework}
\end{figure*}

In this section, we further extend the capability of our model by defining an \textit{Subject-dependent Deep Aging Policy Network} to provide a \textit{planning aging path} for the aging controller. Consequently, the synthesized sequence, i.e. $\{\mathbf{x}^1, \ldots, \mathbf{x}^T\}$, is guaranteed to be the best choice in the face aging development for a given subject.

Let $\zeta_i = \{\mathbf{x}^1_i, \mathbf{a}^1_i, \ldots, \mathbf{x}^T_i\}$ be the observed age sequence of the $i$-th subject and $\zeta = \{\zeta_1, \zeta_2, \ldots, \zeta_M\}$ be the set of all $M$ aging sequences in the dataset. The probability of a sequence $\zeta_i$ can be defined as 
\begin{equation} \label{eqn:SequenceObj}
P(\zeta_i) = \frac{1}{Z} \exp(-E_\Gamma(\zeta_i))
\end{equation}
where $E_\Gamma(\zeta_i)$ is an energy function parameterized by $\Gamma$, and $Z=\sum_{\bar{\zeta}} \exp(-E_\Gamma(\bar{\zeta}))$ is the partition function computed using all possible aging sequences $\bar{\zeta}$.
Then, the goal is to learn a model such that the log-likelihood $\mathcal{L} (\zeta;\Gamma)$ of the observed aging sequences is maximized as follows:
\small
\begin{equation} 
\label{eqn::Log_likelihood_sequence}
\begin{split}
\Gamma^* &= \arg \max_\Gamma \mathcal{L} (\zeta;\Gamma)=\frac{1}{M} \log \prod_{\zeta_i \in \zeta} P(\zeta_i)
\end{split}
\end{equation}
\normalsize
In Eqn. \eqref{eqn::Log_likelihood_sequence}, if $E_\Gamma(\zeta_i)$ is considered as a form of a reward function, then the problem is equivalent to learning a policy network from a Reinforcement Learning (RL) system given a set of demonstrations $\zeta$.

The reward function is the key element for policy learning in RL. However, \textbf{pre-defining a reasonable reward function for face aging synthesis is impossible in practice}. 
Indeed, it is very hard to measure the goodness of the age-progressed images even the ground-truth faces of the subject at these ages are available. 
Therefore, rather than pre-define an add-hoc aging reward, the energy $E_\Gamma(\zeta_i)$ is represented as a neural network with parameters $\Gamma$ and adopted as a non-linear cost function of an \textit{Inverse Reinforcement Learning} problem.

In this IRL system, $\Gamma$ can be directly learned from the set of observed aging sequences $\zeta$. Fig. \ref{fig:AgeProgressionFramework} illustrates the structure of the proposed IRL framework. Based on this structure, given a set of aging sequences as demonstrations, not only the cost function can be learned to maximize the log-likelihood of observed age sequence but also the policy, i.e. predicting aging path for each individual, is obtained with respect to the optimized cost.

Mathematically, the IRL based age progression procedure can be formulated as follows. Let $ \mathcal{M}_{irl} \left\lbrace \mathcal{S}, \mathcal{A}, \mathcal{T}, \zeta, E_{\Gamma} \right\rbrace $ be a Markov Decision Process (MDP) where $\{\mathcal{S}, \mathcal{A}, \mathcal{T}\}$ denote the state space, the action space, and the transition model, respectively. $\zeta$ is the set of observed aging sequences and $E_{\Gamma}$ represents the cost function. Given an MDP 
$ \mathcal{M}_{irl}$, our goal is to discover the unknown cost function $E_{\Gamma}$ from the observation $\zeta$ as well as simultaneously extract the policy $q(\mathbf{a}^{t-1} | \mathbf{x}^{t-1})$ that minimizes the learned cost function.

\textbf{State}: The state $\mathbf{s}^t = \left[\mathbf{x}^t, age\right]$ is defined as a composition of two information, i.e. 
the face image $\mathbf{x}^t$ at the $t$-th stage; 
and the age label of $\mathbf{x}^t$.  

\textbf{Action}: Similar to the age controller, an action  $\mathbf{a}^t$  is defined as the amount of aging variations that the progression process should perform on state $\mathbf{s}^t$. Given $\mathbf{s}^t$, an action $\mathbf{a}^t$ is selected by stochastically sampling from the action probability distribution. During testing, given the current state, the action with the highest probability is chosen for synthesizing process.
Due to data availability where the largest aging distance between the starting and ending images of a sequence is 15, we choose the length of $N_a = 16$ (i.e. plus one state of $\mathbf{a}^t$ where $\mathbf{s}^t$ and $\mathbf{s}^{t+1}$ has the same age).

\textbf{Cost Function}: The cost function plays a crucial role to guide the whole system to learn the sequential policies to obtain a specific aging path for each subject. 
Getting a state $\mathbf{s}^t$ and $\mathbf{a}^t$ as inputs, the cost function maps them to a value $c_\Gamma(\mathbf{s}^t, \mathbf{a}^t)$. Thus, the cost for the $i$-th aging sequence can be obtained as $E_\Gamma(\zeta_i) = \sum_t c_\Gamma(\textbf{s}^t_i, \textbf{a}^t_i) $.
In order to learn a complex and nonlinear cost formulation, each $c_\Gamma(\mathbf{s}^t, \mathbf{a}^t)$ is approximated by a neural network with two hidden layers of 32 hidden units followed by Rectified Linear Unit (ReLU). 

\textbf{Policy}:
Given the cost function $c_\Gamma(\textbf{s}^t_i, \textbf{a}^t_i)$, the policy is presented as a Gaussian trajectory distribution as follows.
\begin{equation}
q(\zeta_i) = q(\mathbf{s}^1_i) \prod_t q(\mathbf{s}^{t+1}_i|\mathbf{s}^t_i, \mathbf{a}^t_i) q(\mathbf{a}^t_i|\mathbf{s}^t_i)
\end{equation}
Then it is optimized respecting to  the expected cost function $\mathbb{E}_{q(\zeta_i)}[E_\Gamma(\zeta_i)]$.

Given the defined state $\mathbf{s}^t$ and action $\mathbf{a}^t$, the observation aging sequence $\zeta_i$ is redefined as $\zeta_i = \{\mathbf{s}^1_i, \mathbf{a}^1_i, \ldots , \mathbf{s}^T_i\}$.
The log-likelihood $\mathcal{L} (\zeta;\Gamma)$ in Eqn. \eqref{eqn::Log_likelihood_sequence} can be rewritten as,
\small
\begin{equation} \label{eqn::Log_Likelihood_seq_rewrite}
\begin{split}
\mathcal{L} (\zeta;\Gamma) &= -\frac{1}{M} \sum_{\zeta_i \in \zeta} E_\Gamma(\zeta_i) - \log \sum_{\bar{\zeta}} \exp(-E_\Gamma(\bar{\zeta}))
\end{split}
\end{equation}
\normalsize
\begin{algorithm}[t]
	\caption{Subject-Dependent Aging Policy Learning}
	\begin{algorithmic}[1]
		\STATEx \textbf{Input:} Observed $M$ age sequence $\zeta=\{\zeta_1,\ldots, \zeta_M\}$ where $\zeta_i = \{\mathbf{s}^1_i, \mathbf{a}^1_i,\ldots,\mathbf{s}^T_i\}$. 
		\STATEx { \textbf{Output:} optimized cost params $\Gamma$ and distribution $q(\zeta)$}
		\STATE \textbf{Initialization:} Randomly initialize policy distribution $q^1_\Gamma(\zeta)$ with a uniform distribution.  
		\FOR { $k = 1$ to $K_1$}	
		\STATE Sample $M$ aging paths from $q^k_\Gamma(\zeta)$.
		\FOR {$i=1$ to $M$}
		\STATE Apply synthesis component in Section \ref{subsec:aging_model} given 
		\STATEx \hspace{0.6cm} the $i$-th sampled aging path and the starting state 
		\STATEx \hspace{0.6cm} $\mathbf{s}^1_i$ to obtain the sampled sequence $\bar{\zeta}_i$. 
		\STATE Add $\bar{\zeta}_i$ to $\zeta_{q^k_\Gamma}$ 
		\ENDFOR
		\FOR { $i = 1$ to $K_2$} 		
		\STATE Sample a batch of observed sequence $\hat{\zeta} \subset \zeta$ 
		\STATE Sample a batch of sampling sequence $\hat{\zeta}_{q^k_\Gamma} \subset \zeta_{q^k_\Gamma}$ 
		\STATE $\zeta_q \leftarrow \hat{\zeta} \cup \hat{\zeta}_{q^k_\Gamma} $
		\STATE Compute the gradient $\frac{d \mathcal{L}( \Gamma)}{d\Gamma} $ using Eqn. \eqref{eqn::gradient_loss}
		\STATE{Update  $\Gamma$ using $\frac{d \mathcal{L}( \Gamma)}{d\Gamma} $ }
		\ENDFOR
		\STATE{Update $q^k_\Gamma(\zeta)$ with $\zeta_{q^k_\Gamma}$ and $E_\Gamma$ to $q^{k+1}_\Gamma(\zeta)$ using approach in \citep{levine2014learning}.}
		\ENDFOR
	\end{algorithmic}
	\label{alg:guide_cost_alg}
\end{algorithm}

Since the computation of the partition function is intractable, the sampling-based approach \citep{finn2016guided} is adopted to approximate the second term of $\mathcal{L} (\zeta;\Gamma)$ in Eqn. \eqref{eqn::Log_Likelihood_seq_rewrite} by a set of aging sequences sampled from the distribution $q(\zeta)$.
\small
\begin{equation}\label{eqn::IRL_Cost_Loss} \nonumber
\mathcal{L} (\zeta;\Gamma) \approx - \frac{1}{M} \sum_{\zeta_i \in \zeta} E_\Gamma(\zeta_i) - \log \frac{1}{N}\sum_{\zeta_j \in \zeta_q}\frac {\exp(-E_\Gamma(\zeta_j))} {q(\zeta_j)}
\end{equation}
\normalsize
where $N$ is the number of age sequences sampled from a sampling distribution $q$. Then
the gradient is given by
\small
\begin{equation}
\label{eqn::gradient_loss}
\nabla_{\Gamma}\mathcal{L} = -\frac{1}{M} \sum_{\zeta_i \in \zeta} \frac{d E_\Gamma(\zeta_i)}{d \Gamma} + \frac{1}{Z'} \sum_{\zeta_j \in \zeta_q} w_{\zeta_j} \frac{d E_\Gamma(\zeta_j)}{d \Gamma}
\end{equation}
\normalsize
where $ Z' = \sum_{\zeta_j} w_{\zeta_j}$ and $w_{\zeta_j} = \frac{\exp(-E_\Gamma(\zeta_j))}{q(\zeta_j)}$.

The choice of the distribution $q$ is now critical to the success of the approximation. 
It can be adaptively optimized by first initialized with a uniform distribution and followed an iteratively three-step optimization process: (1) generate a set of aging sequences $\zeta_q$; (2) update the cost function using Eqn. \eqref{eqn::gradient_loss}; and (3) refine the distribution $q$ as in Eqn. \eqref{eqn::q_optimize}. 
\small
\begin{equation} \label{eqn::q_optimize}
q^* = \arg\min_q \mathbb{E}_q \left[c_\Gamma(\zeta)\right] - \mathcal{H}(\zeta)
\end{equation}
\normalsize
To solve Eqn. \eqref{eqn::q_optimize}, we adopt the optimization approach \citep{levine2014learning} that also results in a policy $q(\mathbf{a}^t|\mathbf{s}^{t})$.
The Algorithm \ref{alg:guide_cost_alg} presents the learning procedure in our policy network and cost function parameters.
\begin{figure*}[t]
	\centering \includegraphics[width=2\columnwidth, height=7cm]{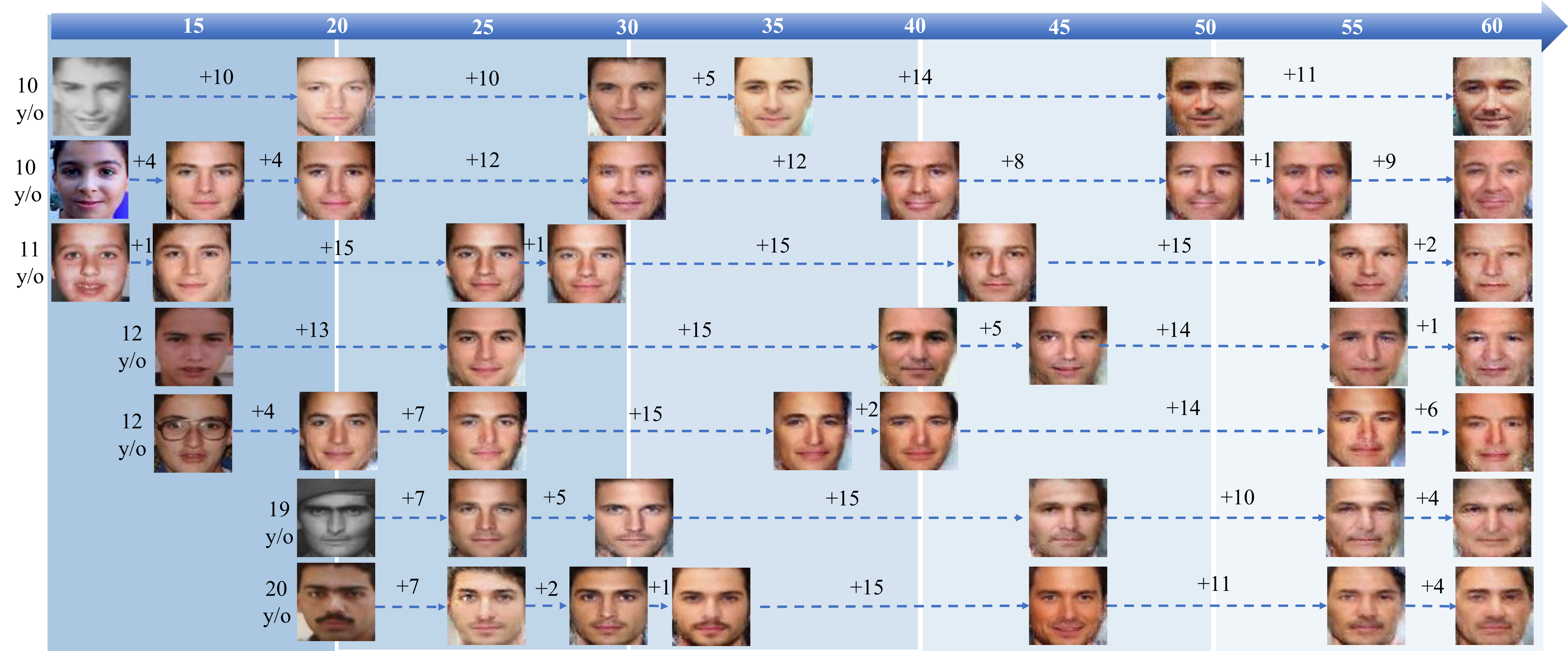}
	\caption{Age progression results on FG-NET. Given images at different ages and the target age of 60s, SDAP automatically predicts the optimal aging path and produce plausible age-progressed faces for each subject. \textbf{Best viewed in color.}}	
	\label{fig:TNVP_AgeProgressedFaces}
\end{figure*} 

\noindent
\textbf{Face aging with single and multiple inputs}: During testing stage, given a face, its inputs, i.e. image and age, are used in the first state $\mathbf{s}^1$. The action for $\mathbf{s}^1$ is predicted by the policy network. Then, the synthesis component can produce the age-progressed face for next state. This step is repeated until the age of the synthesized face reaches the target age.

Using this structure, the framework can be easily extended to take multiple inputs. 
Given $n$ inputs to the framework, they are first ordered by ages and an input sequence $\{\mathbf{s}^1_0, \mathbf{a}_0^1, \ldots, \mathbf{s}^n_0\}$ is created, where $\mathbf{s}^i_0$ denotes the state with the $i$-th input face and age; and $\mathbf{a}^i_0$ is the age difference between $\mathbf{s}^i_0$ and $\mathbf{s}^{i+1}_0$. The synthesis component can be employed to obtain the values for latent variable $\mathbf{z}_n^0$. 
This variable can act as ``memory'' that encodes all information from the inputs. 
We then  initialize $\mathbf{s}^1 = \left[\mathcal{F}_2^{-1}(\mathbf{z}_n^0), age(\mathbf{s}_0^n)\right]$ and start the synthesis process as in the single input case.

\section{Model Properties}
\textbf{\textit{Tractability and Invertibility}.} Similar to its predecessor, i.e. TNVP, with the specific structure of the invertible mapping function $\mathcal{F}$, both inference and generation processes in SDAP are exact, tractable, and invertible. 

\noindent
\textbf{\textit{Generalization}.} During training stage, the action is selected by stochastically sampling from the action probability distribution. This helps our model implicitly handle uncertainty during learning process and is generalized to all the aging steps of age controller.

\noindent
\textbf{\textit{Capability of learning from limited number of face images}.} As we can see in Eqn. (10), the first term is the data-dependent expectation which can be easily computed using training data. For the second term, it is considered as the model expectation and computed via a sampling-based approach. Thanks to the sampling process, our model can still approximate the distribution with a small number of training sequences to be used for the first term.

\section{Discussion}

\begin{figure*}[t]
	\centering 
	\includegraphics[width=1.77\columnwidth]{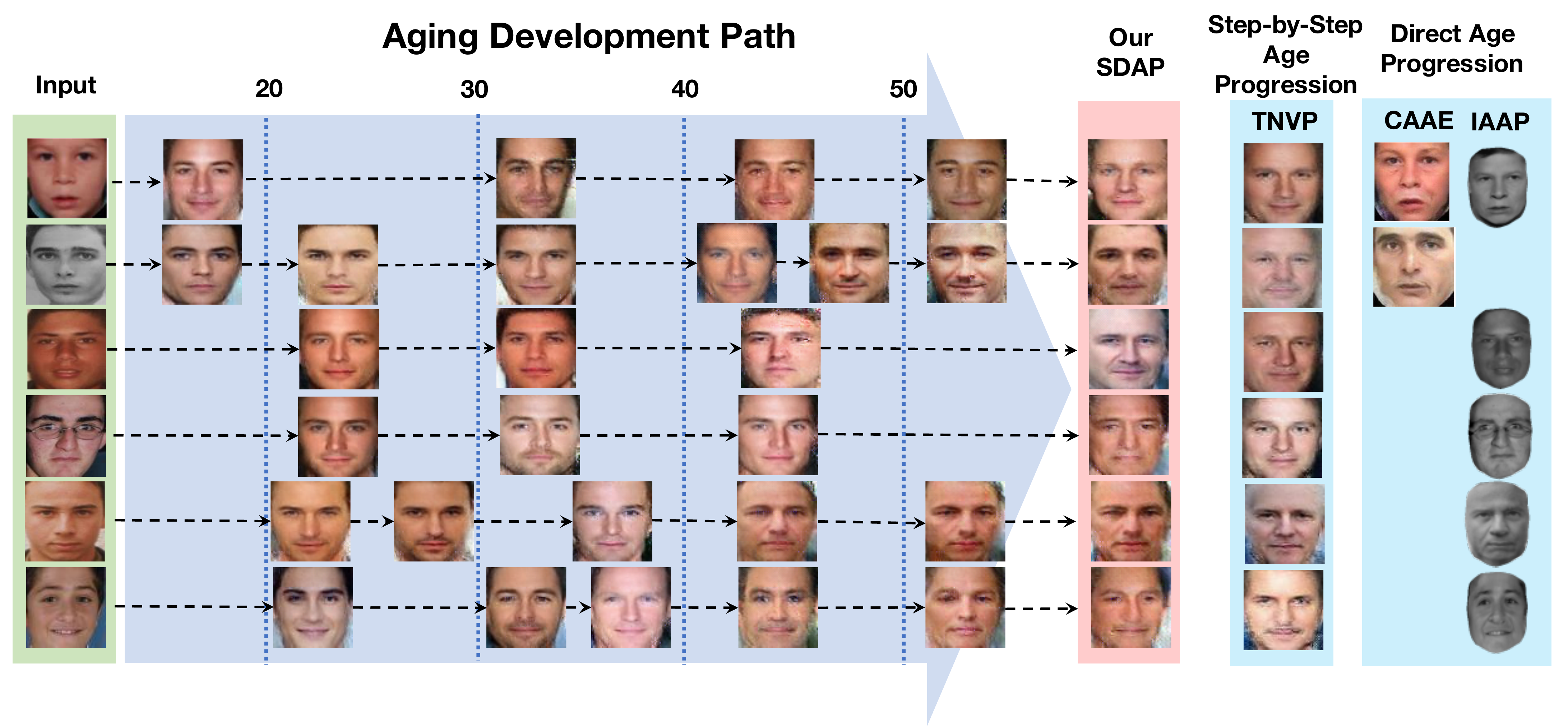}
	\caption{Comparisons between our SDAP against direct approach, i.e. IAAP \citep{kemelmacher2014illumination}, CAAE \citep{Zhang_2017_CVPR}, and step-by-step approach, i.e. TNVP \citep{Duong_2017_ICCV}, on FGNET. \textbf{Best viewed in color.}}	
	\label{fig:TNVP_Comparison}
\end{figure*}

By setting up the invertible mapping functions as deep convolutional networks, SDAP structure is able to shares the advantages of its predecessor, i.e. TNVP, in the capabilities of \textbf{\textit{efficiently capturing highly nonlinear facial features}} while \textbf{\textit{maintaining a tractable log-likelihood density estimation}}. 
Besides aging variation, SDAP is also able to effectively handle other variations such as pose, expression, illumination, etc., as 
can be seen in Figs. \ref{fig:TNVP_AgeProgressedFaces} and \ref{fig:TNVP_Comparison}.

Unlike TNVP, SDAP provides a \textbf{\textit{more advanced architecture that optimizes the amount of aging information to be embedded to input face}}. This ability benefits not only the training process, i.e. maximize the training data usability, but also the testing phase, i.e. flexible and more controlled in the progressed face to be synthesized via the age controller. Fig. \ref{fig:SDAP_DifferentAction} illustrates different age-progressed results obtained by varying the values of the age controller. The bigger gap value produces the older faces.

While previous aging approaches only embed the aging information via the relationships between the input image and age label (i.e. direct approach), or images of two consecutive age groups (i.e. step-by-step) approach, SDAP structure aims at \textbf{\textit{learning from the entire age sequence}} with the learning objective is designed specifically for sequence (see Eqn. \eqref{eqn:SequenceObj}). Under the IRL framework, the whole sequence can be fitted into the learning process for network optimization. As a result, more stable aging sequences can be synthesized. 

Moreover, SDAP's policy learning is more advanced compared to Imitation Learning through supervised learning. In particular, in SDAP, \textbf{\textit{the aging relationship between variables in the whole sequence is explicitly considered}} and optimized during learning process (see Eqn. \ref{eqn::Log_likelihood_sequence}). Therefore, besides the ability of generalization, SDAP is able to \textbf{\textit{recover from ``out-of-track'' results in the middle steps}} during synthesizing. On the other hand, Imitation Learning lacks the generalization capability and cannot recover from failures \citep{attia2018global}. Moreover, since the input face images usually contain different variations (i.e. poses, expressions, etc.) besides age variation, the synthesized results in the middle steps are easily deviated from the optimal trajectory of the demonstration. Consequently, Imitation Learning will produce a cascade of errors and reduce the performance of the whole system.

Compared to direct approach such as GAN-based approach \citep{Yang_2018_CVPR}, our SDAP has several advantages. Firstly, our SDAP \textbf{\textit{provides more control in handling non-linear transformation during aging development process}} since short-term changes can be modeled more efficient. Secondly, different aging features added to different subject's faces are not guaranteed in \citep{Yang_2018_CVPR} (i.e. the same wrinkles can be added to faces of different subjects, and with the presence of the same wrinkles, the age discriminator can be easily fooled). Meanwhile, our SDAP has the capability of discovering the optimal aging development path for each individual and, therefore, \textbf{\textit{producing different aging features for different subjects}}. 
Thirdly, the age discriminator used in this GAN-based model only distinguishes between young and senior age groups, it embeds very coarse relationships between the age label and synthesizing process. On the other hand, in our IRL framework, \textbf{\textit{age label (accurate to year) is also included in a state definition so that finer relationships can be embedded}} during learning process.

\section{Experimental Results}
 
\subsection{Databases}
The proposed SDAP approach is trained and evaluated using two training and two testing databases that are not overlapped. The training sets consist of images from AginG Faces in the Wild \citep{Duong_2016_CVPR} and aging sequences from Cross-Age Celebrity Dataset \citep{chen14cross}. In testing, two common databases, i.e. FG-NET \citep{fgNetData} and MORPH, \citep{ricanek2006morph} are employed.

\textbf{AginG Faces in the Wild (AGFW)}: introduces a large-scale dataset with 18,685 images collected from search engines and mugshot images from public domains.

\textbf{Cross-Age Celebrity Dataset (CACD)} includes 163446 images of 2000 celebrities with the age range of 14 to 62.

\textbf{FG-NET} is a common testing database for both age estimation and synthesis. It includes 1002 images of 82 subjects with the age range is from 0 to 69 years old.

\textbf{MORPH} provides two albums with different scales. The small-scale album consists of 1690 images while the large-scale one includes 55134 images. We use the small-scale album in our evaluation.
\subsection{Implementation Details}

\paragraph{\textbf{Data setting.}} In order to train our SDAP model, we first extract the face regions of all images in AGFW and CACD and align them according to fix positions of two eyes and mouth corners.
Then, we select all possible image pairs (at age $t_1$ and $t_2$) of a subset of 575 subjects from CACD such that $t_1 \leq t_2$ and obtain 13,667 training pairs.
From the images of these subjects, we further construct the observed aging sequence set by ordering all images of each subject by age. This process produces 575 aging sequences.
\paragraph{\textbf{Training Stages.}}Using these training data, we adopt the structure of mapping functions in \citep{Duong_2017_ICCV} for our bijections $\mathcal{F}_1, \mathcal{F}_2$ and pretrain them using all images from AGFW for the capability of face interpretation. 
Then a two-step training process is applied. In the first step, the structure of synthesis unit with two functions $\mathcal{F}_1, \mathcal{F}_2$ and an age controller is employed to learn the aging transformation presented in all  13,667 training pairs. The synthesis units are then composed to formulate the synthesis component.
Then in the second step, the Subject-Dependent Aging Policy Learning is applied to embed the aging relationships of observed face sequences and learn the Policy Network.

\paragraph{\textbf{Model Structure.}} The structure of  $\mathcal{F}_1, \mathcal{F}_2$ includes 10 mapping units where each unit is set up with 2 residual CNN blocks with 32 hidden feature maps for its scale and translation components. 
The convolution size is $3 \times 3$. The training batch size is set to 64.
In the IRL model, a fully connected network with two hidden layers is employed to build policy model. Each layer contains 32 neural units followed by a ReLU activation. The input of this policy network is the state defined in Sec. \ref{subsec:IRL_model} with the dimension of 12289. The output of this policy network is the probability for each action ($N_a$ = 16) and tanh activation function is applied to obtain the predicted action. 
To model the reward/cost function, we adopted a regression network with two hidden layers to predict the reward given the state and action.
Each hidden layer consists of 32 units followed by ReLU operator.  

The training time for our models is 24.75 hours in total on a machine with a Core i7-6700@3.4GHz CPU, 64.00 GB RAM and a GPU NVIDIA GTX Titan X. We develop and evaluate our inverreinforcement learning algorithm \ref{alg:guide_cost_alg} and models using the framework (\textit{rllab}) \citep{duan2016benchmarking}.

\subsection{Age Progression}

Since our SDAP is trained using face sequences with age ranging from 10 to 64 years old, it is evaluated using all faces above ten years old in FG-NET and MORPH. Given faces of different subjects, our aging policy can find the optimal aging path to reach the target ages via intermediate age-progressed steps (Fig. \ref{fig:TNVP_AgeProgressedFaces}).
Indeed, SDAP not only produces aging path for each individual, but also well handles in-the-wild facial variations, e.g. poses, expressions, etc.

\begin{figure}[t]
	\centering 
	\includegraphics[width=0.8\columnwidth]{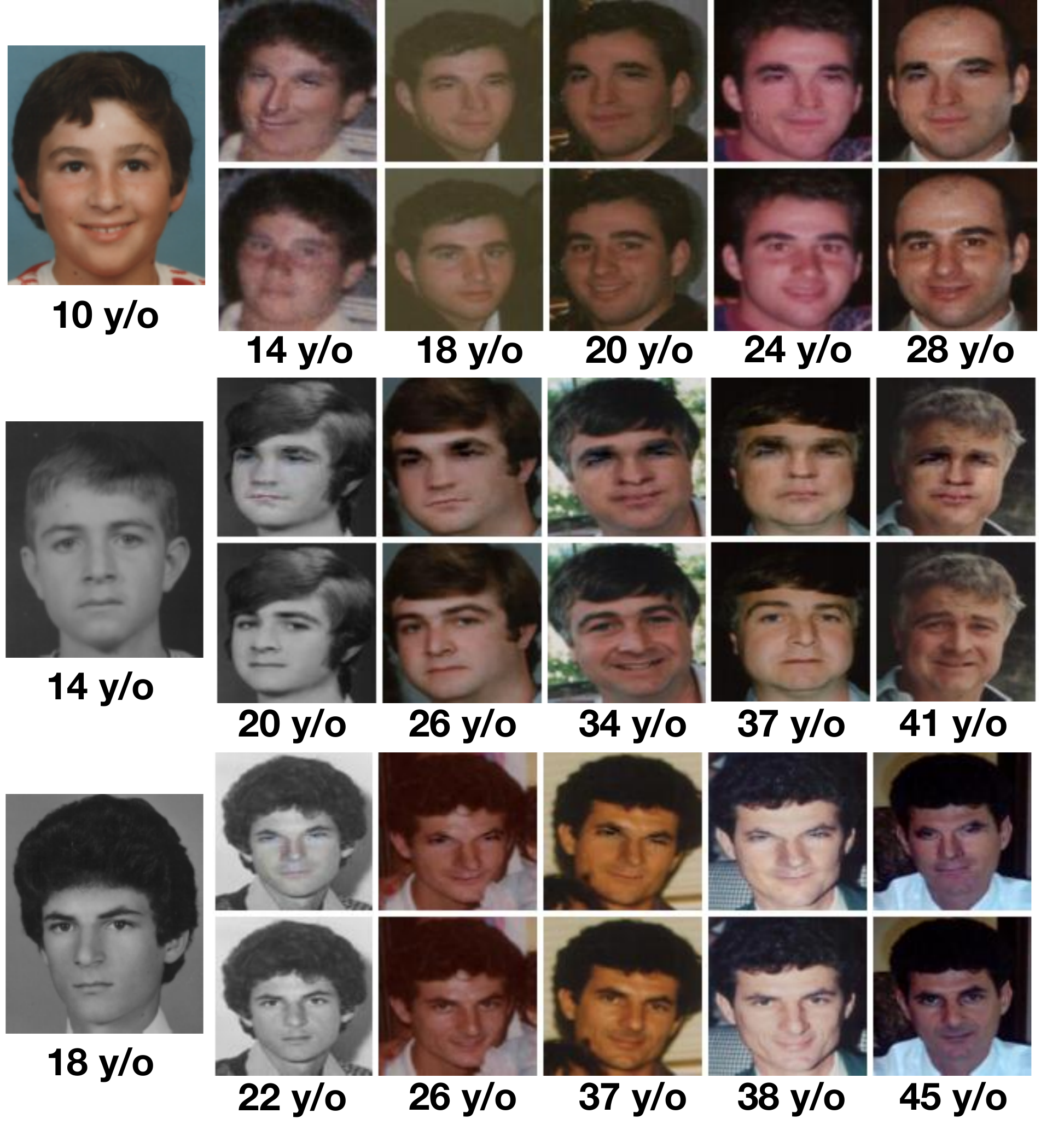}
	\caption{Age progression using SDAP on FGNET. Given images (1st column), SDAP synthesizes the subject's faces at different ages (row above) against the ground-truth (row below). \textbf{Best viewed in color.}}	
	\label{fig:SDAP_Syn_Real}
\end{figure}

\begin{figure}[t]
	\centering 
	\includegraphics[width=1\columnwidth]{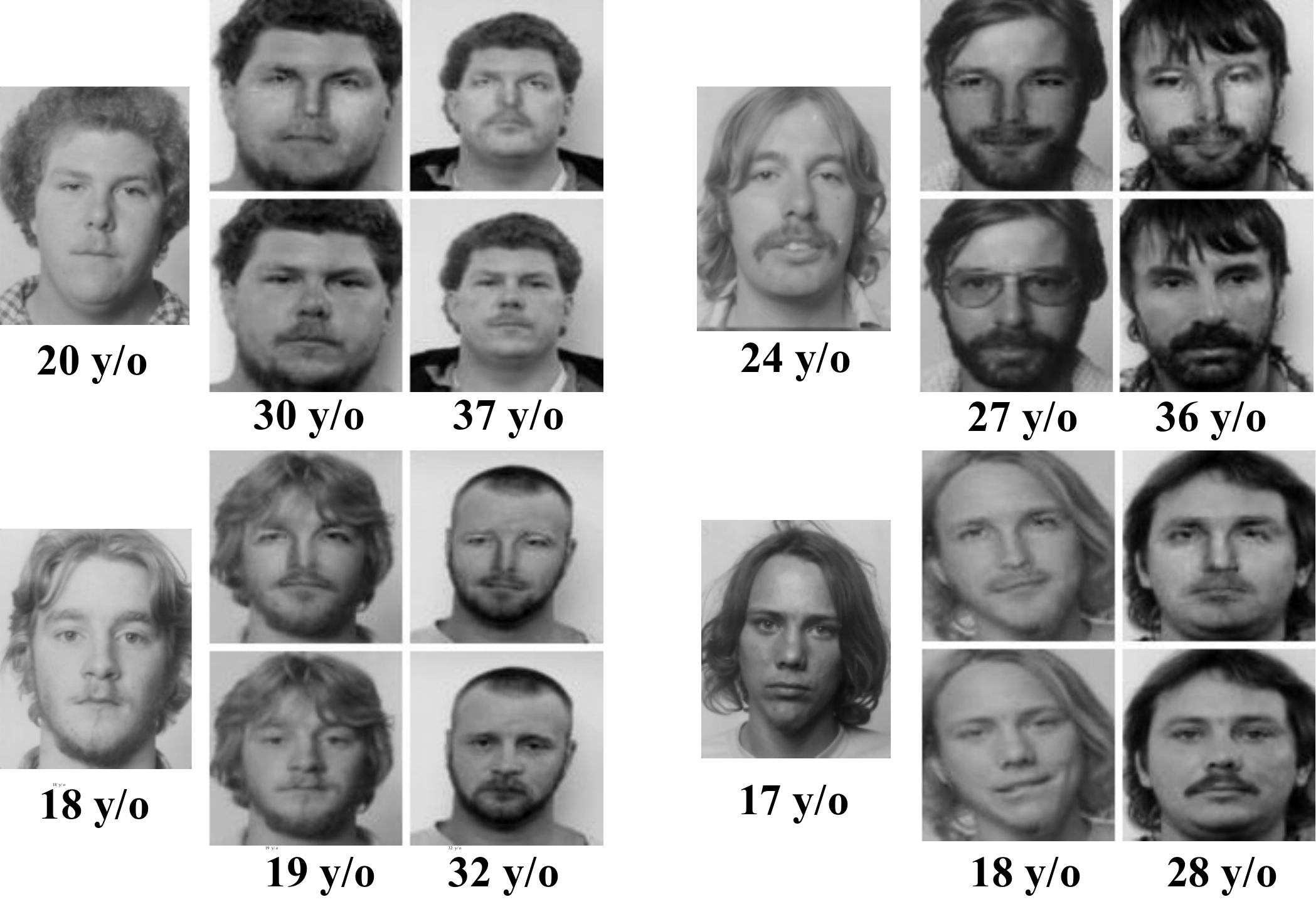}
	\caption{Age progression using SDAP on MORPH. Given images (1st column), SDAP synthesizes the subject's faces at different ages (row above) against the ground-truth (row below). \textbf{Best viewed in color.}}	
	\label{fig:SDAP_Syn_Real_MORPH}
\end{figure}

In addition, the facial textures and shapes are also naturally and implicitly evolved according to individuals differences. 
In particular, more changes are focused around the ages of 20s, 40s and over 50s where beards and wrinkles naturally appear in the age-progressed faces around those ages.
In Fig. \ref{fig:TNVP_Comparison}, we further compare our synthesized results against other recent work, including IAAP \citep{kemelmacher2014illumination}, CAAE \citep{Zhang_2017_CVPR}, and TNVP \citep{Duong_2017_ICCV}. 
The predicted aging path of each subject is also displayed for reference. When the age distance between the input and target ages becomes large, the direct age progression approaches usually produce synthesized faces that are similar to the input faces plus wrinkles. The step-by-step age progression tends to have better synthesis results but still limited in the amount of variations in synthesized faces. SDAP shows the advantages in the capability of capturing and producing more aging variations in faces of the same age group. 
Figs . \ref{fig:SDAP_Syn_Real} and \ref{fig:SDAP_Syn_Real_MORPH} presents our further results at different ages with the real faces as reference.

\subsection{Age Invariant Face Recognition}

Our SDAP is validated using the two testing protocols as in \citep{Duong_2017_ICCV} with two benchmarking sets of cross-age face verification, i.e. small-scale and large-scale sets.

\begin{figure*}[!t]
	\centering 
	\begin{subfigure}{0.6\columnwidth}
		\centering 
		\includegraphics[width=1\columnwidth]{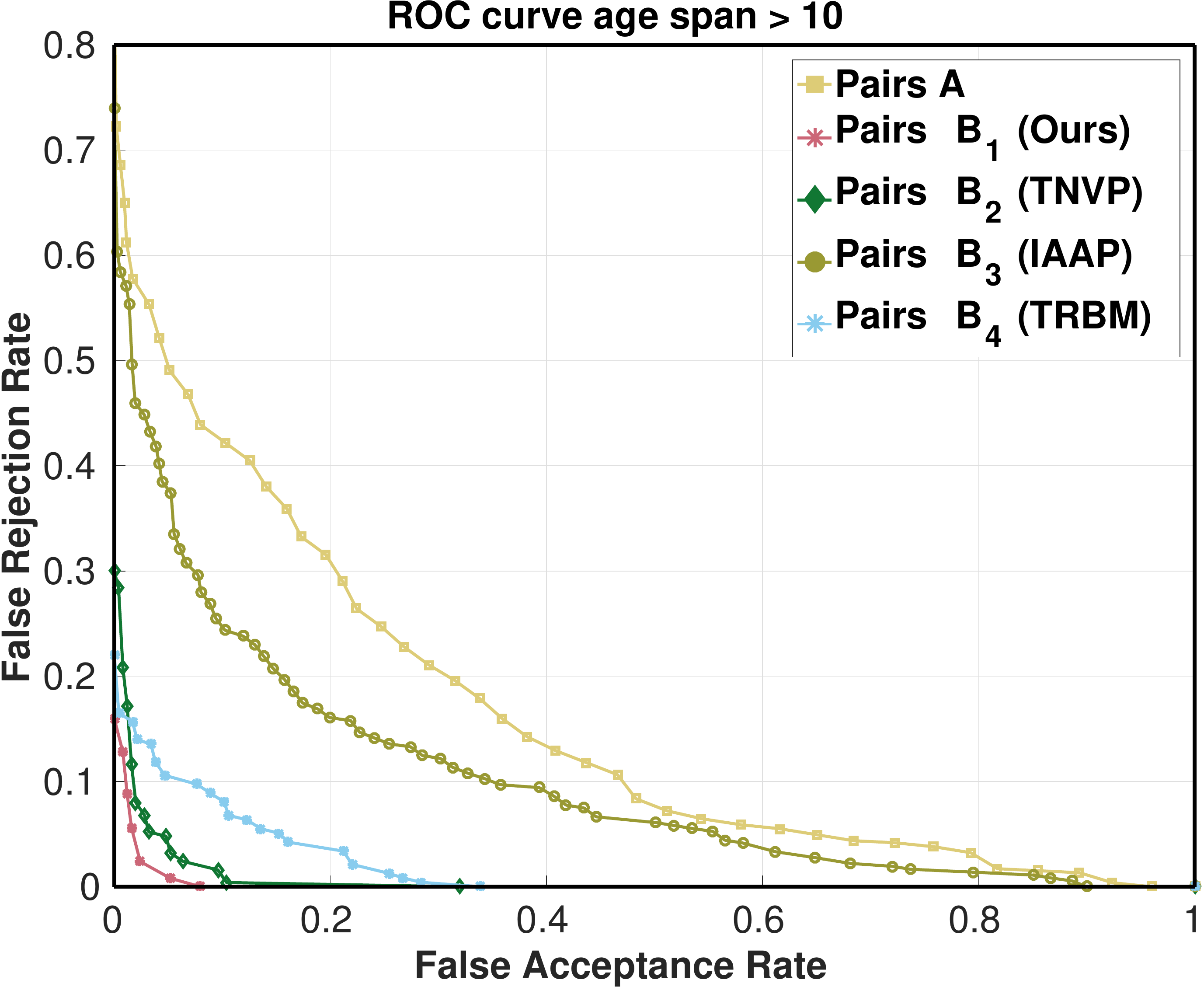}
		\caption{ROC curves of FG-NET pairs}
		\label{fig:FG_ROC_Small}
	\end{subfigure} %
	\begin{subfigure}{0.62\columnwidth}
		\centering 
		\includegraphics[width=0.86\columnwidth]{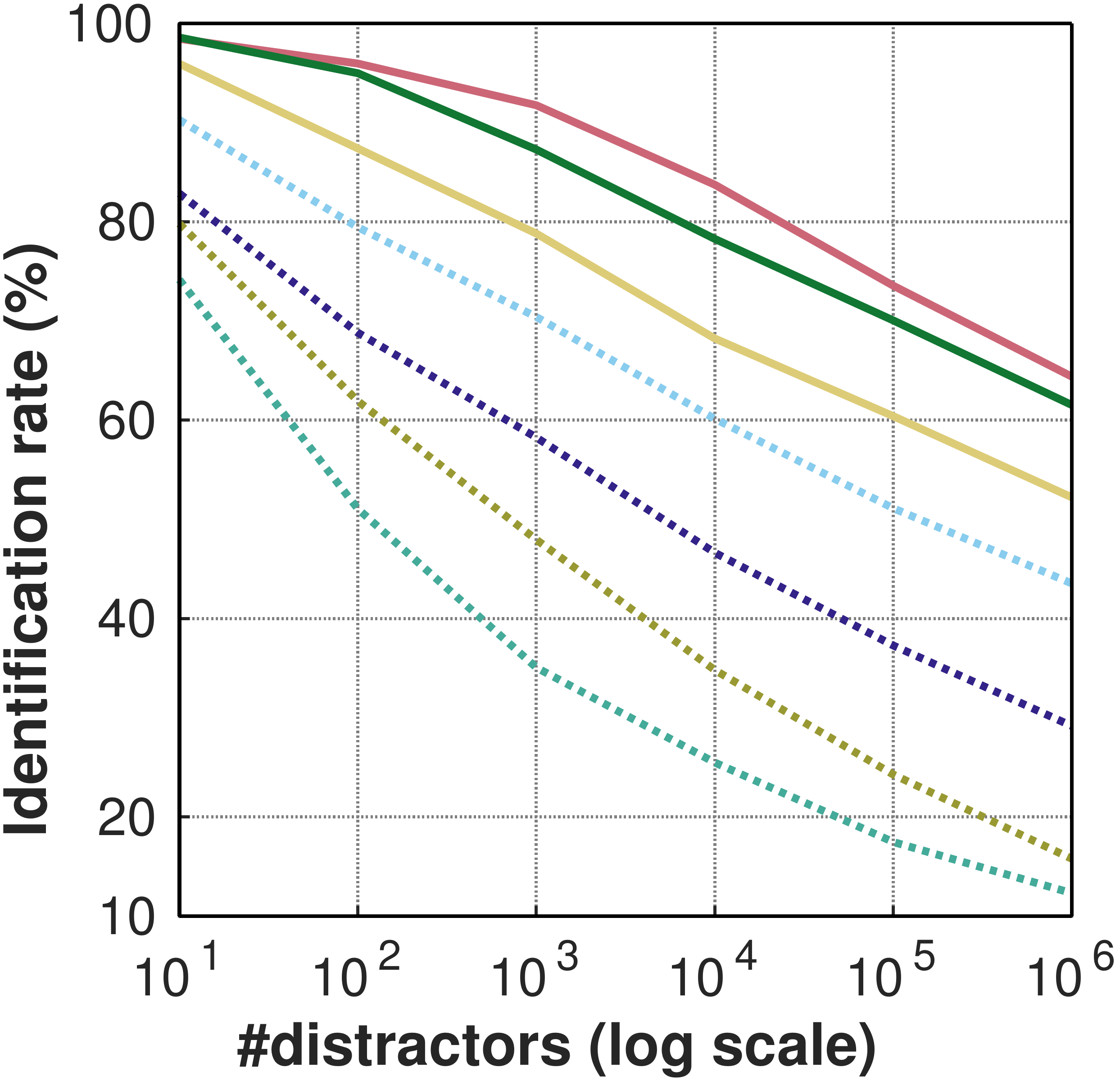}
		\caption{CMC curves on MegaFace}
		\label{fig:megaface_CMC}
	\end{subfigure} %
	\begin{subfigure}{0.8\columnwidth}
		\centering
		\includegraphics[width=1.\columnwidth]{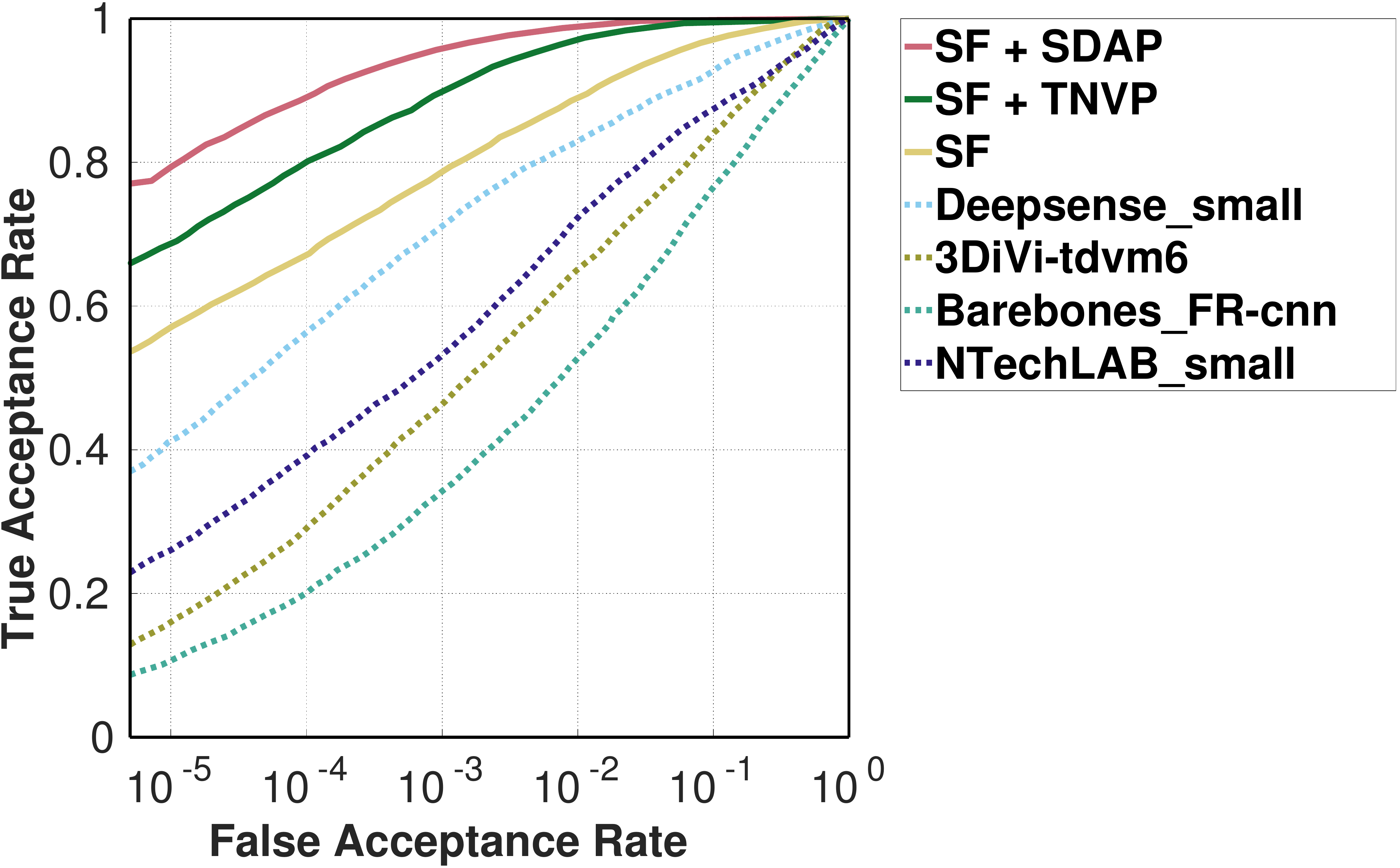}	
		\caption{ROC curves on MegaFace}
		\label{fig:megaface_ROC}
	\end{subfigure} %		
	\caption{  
		Comparison with other approaches in age invariant face recognition (a) ROC curves of face verification on the small-scale protocol; (b) Cumulative Match Curve (CMC) and (c) ROC curves of SF \citep{liu2017sphereface} and its improvements using age-progressed faces from TNVP \citep{Duong_2017_ICCV} and our SDAP on the large-scale protocol of MegaFace challenge 1.}
	\label{fig:megaface}
\end{figure*}

\paragraph{\textbf{Small-scale cross-age face verification.}} In this protocol, we firstly construct a set \textbf{A} of 1052 randomly picked image pairs from FG-NET with age span larger than 10 years old. There are 526 positive pairs (the same subjects) and 526 negative pairs (different subjects).
For each pair in \textbf{A}, SDAP synthesizes the face of the younger age to the face of the older one. This process results in the set $\textbf{B}_1$.
The same process is then applied using other age progression methods, i.e. IAAP \citep{tsai2014human}, TRBM \citep{Duong_2016_CVPR} and TNVP \citep{Duong_2017_ICCV} to construct $\textbf{B}_2$, $\textbf{B}_3$ and $\textbf{B}_4$, respectively. Then, the False Rejection Rate-False Acceptance Rate (FRR-FAR) is reported under the Receiver Operating Characteristic (ROC) curves as presented in Fig. \ref{fig:FG_ROC_Small}. 
These results show that with adaptive aging path for each subject, our SDAP outperforms other age progression approaches with a significant improvement rate for matching performance over the original pairs.

\paragraph{\textbf{Large-scale cross-age face verification.}}
In the \textit{large-scale testing protocol}, we conduct Megaface  face verification benchmarking \citep{kemelmacher2016megaface} targeted on FG-NET plus one million distractors to validate the capabilities of SADP. 
This is a very challenging benchmarking protocol which aims at validating face recognition algorithms not only for the age changing factors but also at the million scale of distractors (i.e. people who are not in the testing set). With this experiment, our goal is to show that using SADP, \textit{the performance of a face recognition algorithm can be boosted significantly without retraining with cross-age databases}.
In this benchmarking, there are two datasets in Megaface including probe and gallery sets. The probe set is the FGNET while the gallery set consists of more than 1 million photos of 690K subjects. Practical face recognition models should achieve high performance against having gallery set of millions of distractors.

Fig. \ref{fig:megaface_CMC} illustrates how the Rank-1 identification rates change when the number of distractors increases. The corresponding rates of all comparing methods at one million distractors are shown in Table \ref{tb:megaface_test}.
Fig. \ref{fig:megaface_ROC} presents the ROC curves respecting to True and False Acceptance Rates (TAR-FAR) \footnote{The results of other methods are provided in MegaFace website.}. The Sphere Face (SF) model \citep{liu2017sphereface}, trained solely on a small scale CASIA dataset having $ <0.49$M images without cross-age information, achieves the best performance among all compared face matching approaches. Using our SDAP aging model, this face matching model can achieve even higher matching results in face verification. 
Moreover, these significant improvements gain without re-training the SF model and it outperforms other models as shown in Table \ref{tb:megaface_test}.

\begin{table}[!t]
	\small
	\centering
	\caption{Comparison results in Rank-1 Identification Accuracy with one million Distractors on MegaFace \#1 FG-NET.} 
	\label{tb:megaface_test} 
	\begin{tabular}{|l|c|c|}
		\hline
		\textbf{Methods} & \textbf{Training set} & \textbf{Accuracy}\\  \hline \hline
		Barebones\_FR & with cross-age faces & 7.136 \% \\
		3DiVi & with cross-age faces & 15.78 \% \\
		NTechLAB & with cross-age faces & 29.168 \% \\
		DeepSense & with cross-age faces & 43.54 \% \\
		\hline \hline
		SF \citep{liu2017sphereface} & without cross-age faces & 52.22\% \\
		SF + TNVP & without cross-age faces & 61.53\% \\
		\textbf{SF + SDAP} & without cross-age faces  & \textbf{64.4}\% \\
		\hline
		
	\end{tabular}
\end{table}

\subsection{Age perceived of synthesized faces}
In this section, the performance of SDAP is further evaluated by assessing the age perceived of the synthesized faces. The goal of this experiment is to validate whether the age-progressed faces are perceived to be at the target ages.
In particular, we adopt the age estimator of \citep{Rothe-IJCV-2016} (i.e. the winner of the Looking At People (LAP) challenge) into the Leave-One-Person-Out (LOPO) protocol.
Then, we compare the Mean Absolute Error (MAE) on real faces and synthesized faces.
In this evaluation, in each fold, all real faces of a subject are selected to construct the testing Set A while the real faces of remaining subjects in FG-NET are used for training. 
Then, for each facial image of the subject in Set A, SDAP is adopted to progress that face to the ages where the subject's real faces are available. This process results in the Set B. Similar processes are also adopted using TNVP \citep{Duong_2017_ICCV} and TRBM \citep{Duong_2016_CVPR} to obtain the Sets C and D, respectively. We repeat this process for all subjects in FG-NET and the age accuracy in terms of MAEs of these sets is showed in Table \ref{tb:MAEs_AgeEstimation}. These results again show that the MAE achieved by SDAP's synthesized faces is comparable to the real faces. Moreover, comparing to the TRBM and TNVP, the difference in MAE between SDAP and real faces are smaller. This further shows SDAP outperforms these approaches in generating the age-progressed faces at the target ages.
Notice that in all experiments, we use only the real faces to train the age estimator. Therefore, these results can confirm the resemblance between the distributions of synthesized and real face images.

\begin{table}[!t]
	\small
	\centering
	\caption{MAEs (years) of Age Estimation System on Real and Age-progressed faces.} 
	\label{tb:MAEs_AgeEstimation} 
	\begin{tabular}{|l|c|}
		\hline
		\textbf{Inputs} & \textbf{MAEs}\\ \hline \hline
		Real Faces (Set A) & 3.10 \\
        \hline \hline
		SDAP's synthesized faces (Set B) & \textbf{3.94} \\
        TNVP's synthesized faces (Set C) & 4.08 \\
        TRBM's synthesized faces (Set D) & 4.27 \\
		\hline
	\end{tabular}
\end{table}

\section{Conclusions}
This work has presented a novel Generative Probabilistic Modeling under an IRL approach to age progression. The model inherits the strengths of both probabilistic graphical model and recent advances of deep network. Using the proposed tractable log-likelihood objective functions together deep features, our SDAP produces sharpened and enhanced skin texture age-progressed faces. In addition, the proposed SDAP aims at providing a subject-dependent aging path with the optimal reward. Furthermore, it makes full advantage of input source by allowing using multiple images to optimize aging path. The experimental results conducted on various databases including large-scale Megaface have proven the robustness and effectiveness of the proposed SDAP model on both face aging synthesis and cross-age verification.

% BibTeX users please use one of
\bibliographystyle{spbasic}      % basic style, author-year citations
\bibliography{egbib.bib}   % name your BibTeX data base

% Non-BibTeX users please use
%\begin{thebibliography}{}
%%
%% and use \bibitem to create references. Consult the Instructions
%% for authors for reference list style.
%%
%\bibitem{RefJ}
%% Format for Journal Reference
%Author, Article title, Journal, Volume, page numbers (year)
%% Format for books
%\bibitem{RefB}
%Author, Book title, page numbers. Publisher, place (year)
%% etc
%\end{thebibliography}

\end{document}